\documentclass[journal=jacsat,manuscript=article]{achemso}

\usepackage[version=3]{mhchem} 
\usepackage{amsmath,amsfonts,amssymb}
\usepackage{tabularx}
\usepackage{booktabs} 
\usepackage{adjustbox}
\usepackage{hyperref}
\usepackage{siunitx}
\usepackage{xcolor}
\usepackage[linesnumbered,ruled,vlined]{algorithm2e}

\author{Hongshuo Huang}
\affiliation[matsci]
{Department of Material Science and Engineering, Carnegie Mellon University, Pittsburgh PA, USA 15213}
\author{Rishikesh Magar}
\affiliation[meche]
{Department of Mechanical Engineering, Carnegie Mellon University, Pittsburgh PA, USA 15213}
\author{Changwen Xu}
\affiliation[meche]
{Department of Mechanical Engineering, Carnegie Mellon University, Pittsburgh PA, USA 15213}
\author{Amir Barati Farimani}
\email{barati@cmu.edu}
\affiliation[meche]
{Department of Mechanical Engineering, Carnegie Mellon University, Pittsburgh PA, USA 15213}
\alsoaffiliation[matsci]
{Department of Material Science and Engineering, Carnegie Mellon University, Pittsburgh PA, USA 15213}
\title[An \textsf{achemso} demo]
  {Materials Informatics Transformer: A Language Model for Interpretable Materials Properties Prediction}

\abbreviations{IR,NMR,UV}
\keywords{American Chemical Society, \LaTeX}

\begin{document}


\begin{abstract}
Recently, the remarkable capabilities of large language models (LLMs) have been illustrated across a variety of research domains such as natural language processing, computer vision, and molecular modeling. We extend this paradigm by utilizing LLMs for material property prediction by introducing our model Materials Informatics Transformer (MatInFormer). Specifically, we introduce a novel approach that involves learning the grammar of crystallography through the tokenization of pertinent space group information. We further illustrate the adaptability of MatInFormer by incorporating task-specific data pertaining to Metal-Organic Frameworks (MOFs). Through attention visualization, we uncover the key features that the model prioritizes during property prediction. The effectiveness of our proposed model is empirically validated across 14 distinct datasets, hereby underscoring its potential for high throughput screening through accurate material property prediction.

\end{abstract}

\section{Introduction}

Machine Learning (ML) models have made significant progress in the field of computational material science\cite{schmidt2019recent,pilania2021machine,keith2021combining,choudhary2022recent}. Their applications span from accurate material property prediction\cite{xie2018crystal,karamad2020orbital,schutt2018schnet,magar2021auglichem,magar2022crystal,cao2023moformer,2020GATGNN,chen2019graph,choudhary2021atomistic,ock2023beyond,kazeev2023sparse} to novel material generation\cite{kim2020generative,xie2021crystal,pollice2021data,wang2022deep,yao2021inverse,ryan2018crystal}. The integration of ML into materials science provides a possible alternative to traditional computational methods, such as Density Functional Theory (DFT) as they are faster surrogates and can facilitates their adoption as high-throughput screening tools\cite{materialsproject}. Notably, Graph Neural Networks (GNNs) stand out as some of the most successful methodologies for material property prediction\cite{fung2021benchmarking}. Owing to their capability to represent local environments and model atomic interactions, GNNs have considerably advanced the domain of material property prediction\cite{reiser2022graph}. Recent improvements include leveraging angular information to construct the line graph\cite{choudhary2021atomistic} to capture three-body interactions. The coNGN\cite{ruff2023connectivity} applied nested graphs of atoms, bond angles, and dihedrals. Encoding periodic patterns, Matformer\cite{yan2022periodic} captured the geometric distances between the same atoms in neighboring unit cells. Potnet\cite{lin2023efficient} embedded interatomic potentials, encompassing the Coulomb potential, London dispersion potential, and Pauli repulsion potential.

Despite these considerable strides, GNNs are not without their limitations, GNNs demand relaxed structures as inputs, a requirement that triggers the need for DFT calculations. This requirement escalates computational costs, which grow directly proportional to the size of the crystal, thereby creating a considerable hurdle to scalability for larger systems like Metal-Organic Frameworks(MOFs) or defects properties in supercell crystal\cite{kazeev2023sparse}. Another challenge faced by GNN models lies in their interpretability. While their message-passing mechanisms are adept at delivering insights at the local atom level by modeling interactions between neighbors. However, they often fail to capture global features especially in capturing the crystal system, lattice parameter and periodicity\cite{gong2022examining}.that necessitate an understanding of the overall structural properties in larger crystals. To overcome these challenges, various computational strategies have been explored such as the coordinate free approch in Wren\cite{goodall2022rapid}, or multi model training. One other approach involves the use of structure-agnostic methods, which aim to bypass the need for specific structural data. These models use formulas as inputs, which helps lower computational requirements\cite{goodall2020predicting,ihalage2022formula,goodall2022rapid}. However, they tend to be less accurate and have difficulty distinguishing between the geometries of different crystalline structures. 

One other novel approach is to represent structures as text and use Large Language Models (LLMs), which has emerged as a promising solution in artificial intelligence. LLMs have been successful in a variety of natural language processing tasks and have shown potentials in molecular machine learning tasks\cite{chithrananda2020chemberta,xu2022transpolymer,cao2023moformer,elnaggar2021prottrans,lin2022language,rao2021msa,yüksel2023selformer,nijkamp2022progen2}. The success in predicting properties of proteins and molecules can be credited to the use of straightforward representations like amino acid sequences (FASTA)\cite{lipman1985rapid} representing either nucleotide sequences or amino acid sequences and SMILES\cite{weininger1988smiles} describing the structure of chemical species. Both FASTA and SMILES are text-based representations, making them naturally compatible with LLMs designed to infer and generate text and can be easily integrated into the transformer architecture. Furthermore, LLMs offer a degree of interpretability as the attention heads can be visualized, providing insights into the most informative tokens that the model deems critical for downstream tasks. The success of language model in molecular area indicate a potential avenue for representing crystalline structures in a manner suitable for transformer models. The ability to convert intricate three-dimensional structural information of materials into tokenized sequences could facilitate the application of LLMs in material property prediction. However, there's no comparable comprehensive representation for crystalline materials that can be easily integrated into the transformer architecture. To leverage the benefits of LLMs in this area, it is imperative to create a representation of crystalline materials. Once this representation is developed, the subsequent challenge is of developing tailored pretraining methods in the context of crystalline materials.

Devising appropriate pretraining methods is fundamental when developing LLMs to predict material properties. Pretraining plays a pivotal role in bolstering the effectiveness of deep learning models. In the context of computational material science, this stage is even more critical due to the sheer complexity and diversity of the materials landscape\cite{ramprasad2017machine}. While GNN models have seen approaches such as Node and Edge Prediction, Self-supervised Learning\cite{magar2022crystal,cao2023moformer}, and Graph Contrastive Learning\cite{wang2022improving,wang2022molclr}, but these method is not suitable for LLMs which typically using Masked Language Modelding (MLM).

This leads us to the following questions: 1.) Can we leverage the inherent crystallographic information such as space group as input representation for the LLMs? 2.) Given an appropriate representation, can we develop a transformer architecture that can accurately predict material properties whilst offering a degree of interpretability of input tokens and flexibility to apply? 3.) Can we develop proper pretraining methods for such transformer architecture to learn crystallography and capture a broad landscape of materials space ?

\begin{figure}
    \centering
    \includegraphics[width=\linewidth]{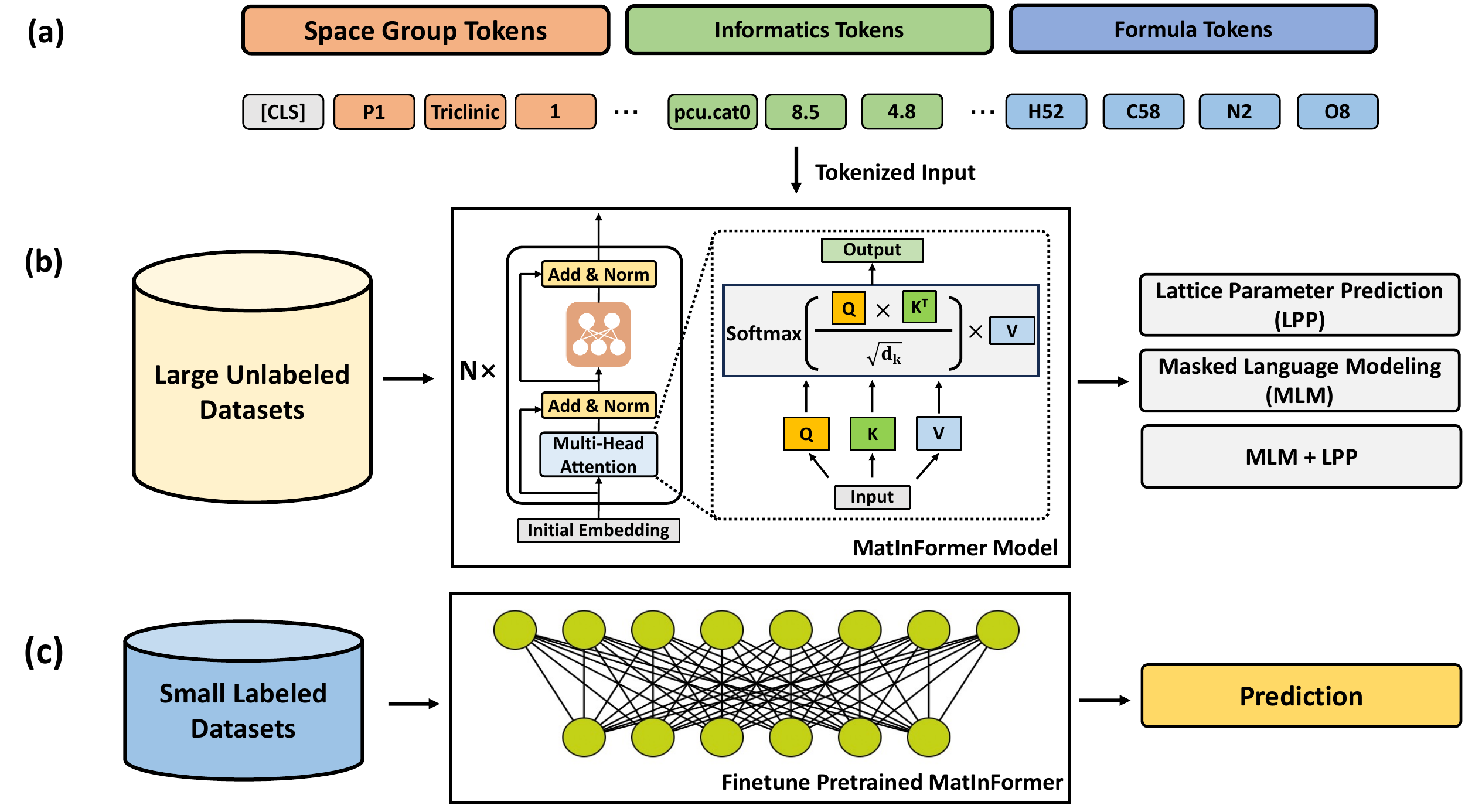}
    \caption{Framework of Material Informatics Transformer. \textbf{(a)}The tokens used for representing material, including Space Group Tokens, Informatics Tokens, and Formula Tokens. The representations are tokenized and combined as the input of the Transformer model. \textbf{(b)} The pretrainning stage. A Large unlabeled dataset  is used to pretrain the transformer encoder with 3 different pretrain method. \textbf{(c)} Fintuning stage for the downstream tasks. The pretrained  transformer is then finetuned for different labeled material property prediction tasks.}
    \label{fig1}
\end{figure}

In this work, we introduce  Materials Informatics Transformer (MatInFormer) that attempts to address some of the aforementioned questions (Figure~\ref{fig1}). We formulate a robust textual representation for crystalline materials, The architecture is capable of learning about the crystalline system by tokenizing the space group information that can in some ways capture the geometry of the crystals, informatics tokens depending on type of crystalline material, and formula tokens that capture the composition of the crystalline material(Figure~\ref{fig1}a). Using the information from the space group, informatics, and formula tokens we feed it into the MatInFormer model based on the Roberta architecture\cite{liu2019roberta}. We pretrain the MatInFormer using three different strategies. For the first strategy, we pretrain the MatInFormer(Figure~\ref{fig1}b) using the classic Masked Language Modeling(MLM). This enables the model to capture the unique characteristics of the crystal system. Additionally, we also pretrain the model predicting the lattice parameters. Finally, we combine the MLM and lattice parameter prediction and use it for pretraining the model. The pretrained model is later finetuned for different downstream tasks(Figure~\ref{fig1}c). Using the MatInFormer architecture we predict the properties of the material for a diverse range of materials from the Matbench suite\cite{dunn2020benchmarking}. Furthermore, we also investigate the efficacy of the model for property prediction of Metal-Organic Frameworks(MOFs)\cite{rosen2021machine,wilmer2012large}. We demonstrate the interpretability aspect of the models by analyzing how different tokens influence the performance of the model. Moreover, we also demonstrate the flexibility of our framework by easily manipulating the number of tokens that are input into the transformer model and illustrating how they influence the performance of the model. Our model MatInFormer bridges LLMs with computational material science, presenting an interpretable framework for property prediction. 


\section{Methods}
\subsection{Materials Informatics Tokenization}

The initial input to the MatInFormer model is an embedded representation produced through the tokenization of available material information. This tokenization process for the MatInFormer model divides the material representation into three distinct parts: space group tokens, informatics tokens, and formula tokens. 

The first of these is the space group tokens. In the field of crystallography, a crystalline material structure is specified by a combination of the structure's space group, the dimension of its unit cell, and the element at its Wyckoff position\cite{hiller1986crystallography, wyckoff1922analytical}. To employ a coordinate-free approach, we discard precise atomic positions, lattice parameters, and Wyckoff position, retaining only the space group as the input. There are 230 space groups in total, which combine 32 crystallography point groups with 14 Bravais lattices. We have standardized the sequence length for space group tokens to 12, as shown in Table~\ref{space group token}. The space group is mapped using a dictionary to ensure a comprehensive representation of group-specific information. This includes the space group symbol, point group, crystal system, and their corresponding group properties. For instance, the space group tokens for $Fm\bar{3}m$ (No. 225) are [$F\frac{4}{m}\bar{3}\frac{2}{m}$],[225],[192],[$m\bar{3}m$],[cubic],[$m\bar{3}m$],[Centrosymmetric],[non-polar],[F],[$\frac{4}{m}$],[$\bar{3}$],[$\frac{2}{m}$]. The space group token representation follows the exact order as specified in Table~\ref{space group token}.  
We calculated the space group symbol using Pymatgen's SpacegroupAnalyzer. This analyzer takes threshold distance and angle tolerance as parameters to determine the Space Group. To determine the space group, we used the default settings with a threshold distance of $0.01 \AA$ and and an angle tolerance of $5.0^{\circ}$, respectively. We would like to note, that these default settings in Pymatgen are slightly different from those in the Materials Project\cite{materialsproject}, where the tolerance for distance is set at $0.1\AA$. In real material systems, there's often a degree of off-site relaxation from the high-symmetry sites, leading to some materials being classified as "P1". However, this distance tolerance only marginally impacts the model's performance on downstream tasks, which aligns with findings from previous studies, such as Wren\cite{goodall2022rapid}.

\begin{table}[htb!]
    \centering
    \begin{tabular}{l|l}
    \toprule
    Token Index & Token\\
    \midrule
    
    0& Full Space Group Symbol  \\
    1& Space Group Number      \\
    2&Order \\
    3&Point Group  \\
    4&Crystal System  \\
    5&Laue Class  \\
    6&Symmetry  \\
    7&Polar \\
    8&Separate Space Group Token 0 \\
    9&Separate Space Group Token 1 \\
    10&Separate Space Group Token 2 \\
    11&Separate Space Group Token 3 \\
    \midrule
    12&Topology\\
    13&Unit Cell Volume\\
    14&Atoms Number\\
    15&Porosity Fraction\\
    16&Accessible Void Fraction\\
    \bottomrule
    
    \end{tabular}
    \caption{Space group and Informatics Tokens for MatInFormer}
    \label{space group token}
\end{table}

The second component of the tokenized input involves informatics tokens. These capture essential attributes associated with a particular type of crystalline system. Consider, for instance, Metal-Organic Frameworks (MOFs). To distinguish between different structures within the hypothetical MOF (hMOF) dataset, relying solely on the formula and space group proves insufficient. More than 53\% of the samples in the hMOF database have duplicate formula and space group. If we use only the space group and formula tokens, the model will determine these to be the same materials. However, those materials have different 3D structure which leads to them having a  different labels. As such, we turn to other tokens that can provide more information that distinguishes the MOF in question. The first such token is topology, drawn from the MOFid as per Bucior et al. (2019)\cite{bucior2019identification}. Every MOFid includes topology and catenation codes adopted from the Reticular Chemistry Structure Resource (RCSR) database\cite{o2008reticular}. In addition to topology, we tokenize the unit cell volume and the number of atoms. This allows the model to potentially learn porosity features through the formula and volume. To guarantee the model's capacity to capture these porosity features, we tokenize the porosity fraction and the accessible void fraction as calculated by porE\cite{trepte2021pore}. To calculate the porosity we apply the Grid Point Approach (GPA). According to GPA, a numerical grid near any atom (within the van der Waals sphere) is considered occupied, while any other grid is seen as unoccupied. Consequently, the void fraction is calculated using Equation~\ref{void}:

\begin{equation}
\Phi_{void} = \frac{N_{unoccupied}}{N_{total}} \cdot 100 \%
\label{void}
\end{equation}
where $\Phi_{void}$ represents the void fraction, $N_{unoccupied}$ represents the number of unoccupied grids, and $N_{total}$ signifies the total number of grids. The accessible porosity is estimated under the premise that a sphere with a probe radius ($r_{probe}$) does not interact with the atoms. Additional details about the GPA approach that we used for porosity calculation are available in SI(Figure S5).

Taking hybrid organic-inorganic perovskites (HOIPs) as another example, we treat the organic cation as an informatics token. The application of informatics tokens lends a level of flexibility to the architecture. This flexibility can be used to incorporate additional information and simultaneously minimize redundancy.

The third component of our tokenization process is the formula tokens. The stoichiometric formula is used to generate an elemental information embedding using "Matscholar" embedding\cite{tshitoyan2019unsupervised}. The elemental information has dimensionality of $d_{el} = 200$, concatenated with each element in the composition weighted by its respective elemental fraction. Finally, the formula embeddings shape in $(N,201)$ where N represents the number of the formula tokens(also the number of the elements). We standardize the sequence length of the formula tokens to 20, which corresponds to the maximum number of elements observed in our dataset. If a particular stoichiometric formula does not contain 20 elements, we employ zero-padding to achieve a consistent length across all formula tokens.

Finally, to ensure compatibility across all token types, the formula tokens are projected into the appropriate dimension using a single linear layer. Following this transformation, all three types of tokens—formula, space group, and informatics—are simultaneously concatenated to form a unified token set. This aggregated set is then fed into the transformer, offering a comprehensive representation of the crystalline material system.

\subsection{Transformer encoder}

Our MatInFormer is based on a Roberta transformer architecture\cite{liu2019roberta} from Huggingface\cite{wolf2019huggingface} library. The MatInFormer model employs self-attention layers as its primary method of operation using "Scaled Dot-Product Attention". The input data are mapped into 3 different embeddings to calculate the attention: queries(Q), keys(K), and values(V). And the attention score is calculated

\begin{equation}
    Attention(Q,K,V) = softmax(\frac{QK^T}{\sqrt{d_k}})V
    \label{qkv}
\end{equation}
where $d_k$ is the dimensions of $K$.

Materials Informatics Transformer is able to capture bidirectional information meaning it can processes the input data from both left-to-right and right-to-left directions and is able to capture features from both directions. The transformer encoder with 8 blocks, 12 heads, and 768 hidden sizes is used. A [CLS] token is located at the first position in each layer. The [CLS] embeddings of the last layer are used for the downstream task. 

\subsection{Pretraining}

Pretraining significantly enhances the learning capabilities of LLMs, especially when trained on vast unlabeled datasets. Our pretraining datasets are sourced from JARVIS-Tools\cite{choudhary2020joint}, which consists of six distinct datasets\cite{Downs2003}. The first dataset, curated from five out of the six datasets from JARVIS-Tools (excluding the Open Catalyst dataset\cite{chanussot2021open}), contains 642,459 data samples. The second dataset combines all six datasets, including the Open Catalyst dataset, totaling 997,528 samples. The dataset with pretraining size 642,459 is named as pretraining dataset 1 and dataset of size 997,528 is named as pretraining dataset 2.  Additional details about the pretraining datasets are available in Table S1. Moreover, we also explore crystal system distribution(Figure S1 and Figure S3) and elemental distribution(Figure S2 and Figure S4) for both the pretraining datasets. In this work, we have developed three distinct pretraining strategies namely masked language modeling(MLM), Lattice Parameter Prediction(LPP), and a combination of MLM and LPP.

\subsubsection{Masked Language Modeling (MLM)}
Our first strategy leverages the Masked Language Modeling (MLM) objective for pretraining the MatInFormer. We employ a masking ratio of 25\% for space group tokens during pretraining, thereby enabling the model to learn and infer the masked space group information. This process enables the Materials Informatics Transformer to understand the intricate crystallographic relationships among 'Space Group', 'Point Group', and 'Crystal System' tokens. Under the MLM strategy, we task the model with predicting the masked space group tokens, thereby enabling it to learn crystallographic information. This includes the classification of the crystal system based on a given space group, or the prediction of the space group number from individual space group tokens. The model is pretrained using the cross-entropy loss function and trained for 50 s under the MLM objective.  The pretraining hyperparameters for the MLM strategy are given in Table S3.

\subsubsection{Lattice Parameter Prediction (LPP)}
The second strategy involves lattice parameters prediction(LPP) of the unit cell (namely, $a, b, c$ and $\alpha, \beta, \gamma$). This approach empowers the model to understand the crystallography interplay among space groups, formulas, and lattices. Unlike the first strategy of MLM, this technique obliges the model to not only understand crystallography and group symmetry, but also to incorporate formula and elements in order to reconstruct the coordinates. In contrast to masked language modeling, this approach enables MatInFormer to learn via crystal system characterization, relationships between lattice parameters, and also take into account the composition of the material. For example, if the crystal system token indicates a [hexagonal] system, the model needs to predict the lattice parameters adhering to the geometric characteristics of a hexagonal system ($a=b\neq c$, $\alpha=\beta=90^{\circ}$, $\gamma=120^{\circ}$). For the pretraining process, we utilized multi-task prediction, implementing a Multilayer Perceptron (MLP) regression head with six outputs for all the lattice parameters. The pretraining hyperparameters for the LPP strategy are given in Table S3.

\subsubsection{Masked Language Modeling and Lattice Parameter Prediction}
For the third pretraining strategy, the MLM and LPP methods combined together. We mask 25\% of the space group tokens and the model is required to predict the lattice parameter using rest of the tokens. Using such a strategy we are able to capture the relationships between space group tokens in addition to using the available crystal information that has not been masked to predict the lattice parameters. The pretraining hyperparameters for the MLM + LPP strategy are given in Table S3.

\subsubsection{Training Details}

In the pretraining stage, the transformer encoder is coupled with a 2-layer MLP regression head. Depending on the pretraining objective we choose the loss function. For the MLM objective, we use the cross entropy loss function. For lattice parameter prediction, the model is trained to minimize the Mean Square Error(MSE) between prediction and the target lattice parameters.  Additional details about the hyperparameters used for pretraining the model are available in SI.

\subsection{Finetuning}
To evaluate the performance of MatInFormer and ascertain the efficacy of our pretraining strategies, we fine-tuned it on several downstream property prediction tasks. We assessed MatInFormer's performance using the Matbench suite, as well as the HOIP\cite{kim2017hybrid}, hMOF\cite{wilmer2012large} datasets. During fine-tuning, our primary inputs were the "Formula" and "Space Group" tokens. Depending on the specific dataset, we supplemented these with informatics tokens.

When fine-tuning, we utilized the weights from the pretrained transformer encoder. For a comparative analysis, we also benchmarked against a model trained from scratch, that was initialized with random weights. For the final property prediction, we used the [CLS] token coupled with a prediction head consisting of 2 MLP-layers with SiLU\cite{elfwing2018sigmoid} activation.

For the Matbench tasks, we adhered to the standard 5-fold cross-validation protocol. For the hMOF dataset, we maintained a 70\%/15\%/15\% train/validation/test split, mirroring the approach taken by MOFormer. Here, the additional informatics tokens included topology, volume of the unit cell, number of atoms, and porosity. Furthermore, we benchmarked against the HOIP dataset\cite{kim2017hybrid}, employing an 80\%/20\% split for training and testing, respectively. In the case of the HOIP dataset, we introduced the organic group as an extra informatics token. Detailed finetuning hyperparameters\cite{loshchilov2018decoupled,loshchilov2016sgdr} can be found in the supplementary information (SI)(Table S4).

\section{Results and Discussion}


To evaluate the model performance on materials property prediction, we use 8 datasets from the Matbench suite.
Moreover, we also benchmark hMOF datasets about $CO_2$ and $CH_4$ absorption for different pressure settings, shown in Table~\ref{benchmark}. The selected datasets not only cover an extensive spectrum of properties such as exfoliation energy, phdos peak, refractive index, modulus, formation energy, band gap, and gas absorption but also vary from each other a lot in terms of the size of the datasets, ranging from 636 to 132,752. We compare our the performance of our model on Matbench to other structure agnostic models: Roost and Finder; Structured-based GNN models: ALIGNN and CGCNN; Coordinate-free model: Wrenformer. Among the baselines, Wrenformer is primarily used for comparison since we use the same transformer architecture and share the same level of coordinate-free information. And for hMOF, we compare our result to CGCNN, MOFormer, and Stoichiometric-120.  and hMOF result is shown in Table \ref{hMOF}.The results for the standalone HOIP dataset are shown in Table S5.
\begin{table}[htb!]
  \centering
  \small
  \footnotesize
  \resizebox{\textwidth}{!}
  {
  \begin{tabular}{l|llllll}
    \toprule
    Dataset & \# Samples & Property  & Unit \\
    \midrule
    JDFT2D (JDFT)\cite{choudhary2017high} & 636 & Exfoliation Energy & {meV per atom}\\
    Phonons\cite{petretto2018high} & 1,265 &Last Phdos Peak & {1 per cm}\\
    Dielectric\cite{petousis2017high} & 4,764 & Refractive Index & {Unitless} \\
    GVRH\cite{ward2018matminer,deJong2015} & 10,987 & Shear Modulus & {$\log_{10}GPa$}\\
    KVRH\cite{deJong2015} & 10,987 & Bulk Modulus & {$\log_{10}GPa$}\\
    Perovskites\cite{Castelli} & 18,928 & Formation Energy  & {eV per atom}\\
    MP-Gap (MP-BG)\cite{materialsproject} & 106,113 & Band Gap & {eV}\\
    MP-E-Form (MP-FE)\cite{materialsproject} & 132,752 & Formation Energy & {eV per atom} \\
    hMOF \cite{wilmer2012large} & 102,858 & Gas Absorbtion & {$mol*kg^{-1}$}\\
    HOIP \cite{kim2017hybrid} & 1,333 & Band Gap & {eV}\\
    
   \bottomrule
  \end{tabular}
}
  \caption{Overview of the datasets used for benchmarking the performance of the Pretraining framework. We predict the properties on 8 different datasets\cite{choudhary2017high,petretto2018high, kim2017hybrid,PhamOrbital,petousis2017high,ward2018matminer,deJong2015,Castelli,materialsproject} aggregated from the Matbench suite\cite{dunn2020benchmarking} and hMOF.}
  \label{benchmark}
\end{table}
 
\subsection{Matbench}

The results for Matbench are shown in Table \ref{matbench}. We compared the performance of MatInFormer, trained from scratch, with that of Wrenformer. Wrenformer uses a coordinate-free transformer based on the Wyckoff representation. Out of 8 tasks, MatInFormer performed better than Wrenformer in 6. Notably, MatInFormer predicted the phonons dataset with 47.0\% lower MAE than Wrenformer did. Also, MatInFormer showed an average reduction of 11.87\% in MAE across the smaller datasets of jdft2d, phonons, dielectric, and modules. There was a 6.93\% decrease in MAE for the band gap and formation energy datasets. Although MatInFormer has an architecture similar to Wrenformer, it still performed better without using detailed Wyckoff positions and representations. This improvement suggests that our model benefits from the Roberta encoder. Additionally, tokenizing space group tokens to provide crystallography information and symmetry properties directly enhances the model's performance. These results indicate that MatInFormer has some state-of-the-art results for coordinate-free models. When we compare MatInFormer to structure-agnostic models like Roost, the performance improvement is clear. MatInFormer showed improvements of 12.62\% for GVRH, 11.71\% for KVRH, and 23.63\% for formation energy.

We also pretrained the MatInFormer using three different strategies. We observed that when MatInFormer was pretrained using Masked Language Modeling (MLM), its performance declined compared to when trained from scratch for some datasets. During the MLM pretraining, random space group tokens are masked and the model is trained to predict those mask tokens with other space group tokens. For example, the models need to predict the crystal system "Orthorhombic" and point group "mm2" for the given space group symbol "Pmm2". However, formula tokens will stay the same since there is no way to predict the formula for a given space group and result in only crystallography information being captured by the model. We noticed that the MLM objective did not help improve performance greatly on downstream tasks. One possible reason for this might be that the space group tokens mainly focus on the geometric properties of the crystal system, and don't really capture the crystalline materials composition. Given the observed limitations with MLM, we then pretrained the model via Lattice Parameter Prediction (LPP), where the model has to not only learn the information about the rules of the crystal system but also needs to consider the impact of formula tokens. Finally, aiming to leverage the strengths of both approaches, we adopted a hybrid strategy that combined both MLM and LPP for pretraining. 

We found that employing the LPP and MLM + LPP pretraining strategies significantly enhanced the performance of the baseline MatInFormer. For instance, using the LPP strategy led to an average improvement of 5.36\%, with the jdft2d, phonons, and band gap datasets showing notable improvements of 13.28\%, 11.83\%, and 8.09\%, respectively. Similarly, the MLM + LPP strategy yielded an average improvement of 4.45\%, with the maximum performance gains again observed in the jdft2d, phonons, and band gap datasets. However, when we pretrained another model using the LPP strategy on a larger dataset with OCP\cite{chanussot2021open} data, the average improvements reduced to 2.36\%. This decrease could potentially be attributed to the presence of unrelaxed structures in the OCP data, which implies that the lattice parameter values are suboptimal, making pretraining to predict them less benefical.

When comparing the pretrained MatInFormer to structure-based models like CGCNN, MatInFormer outperforms in 6 out of 8 tasks after being pretrained using Lattice Parameter Prediction. Unlike CGCNN, MatInFormer processes only the space group tokens and the formula tokens. This indicates that there is potentially a lot of information for models to tap into, beyond just atom distances, angular data, and the extensive feature engineering employed by many structure-based GNNs. The space group captures the patterns and symmetries in crystallography, which some GNN models may not effectively harness. 

Even though MatInFormer achieves performance comparable to GNNs for some tasks, the performance gap between structure-based GNNs and other models lacking atomic distance and angle information is stark, particularly for Perovskites dataset.. The MAE of our model aligns with that of structure-agnostic models like Roost. MatInFormer's weaker performance suggests that perovskite formation energy might depend more on detailed structural attributes, such as atomic positions, than on just formulas or elements. While Wrenformer slightly outperforms MatInFormer by incorporating additional Wyckoff position embeddings, the MAE for these models is still notably higher than that of structure-based GNNs

\begin{table}[h!]
    \centering
    \resizebox{\textwidth}{!}{
    \begin{tabular}{l|llllllll|l}
    \toprule
    Models & jdft2d & phonons & dielectric & GVRH & KVRH & Perovskites & Band Gap & E form& Type\\
    \midrule
    ALIGNN & 43.42$\pm$8.94 & 29.53$\pm$2.11 & 0.3449$\pm$0.0871 & 0.072$\pm$0.001 & 0.057$\pm$0.003 & 0.029$\pm$0.001 & 0.186$\pm$0.003 & 0.022$\pm$0.001 & Structure-based\\
    CGCNN  & 49.24$\pm$11.58 & 57.76$\pm$12.31 & 0.599$\pm$0.083 & 0.090$\pm$0.002 & 0.071$\pm$0.003 & 0.045$\pm$0.001 & 0.297$\pm$0.004 & 0.034$\pm$0.001 & Structure-based\\
    \midrule
MatInFormer &49.61$\pm$14.02 & 48.85$\pm$8.71 & 0.327$\pm$0.083 & 0.090$\pm$0.001 & 0.070$\pm$0.003 & 0.407$\pm$0.008 & 0.290$\pm$0.005 & 0.065$\pm$0.001 & Coordinate free\\
MatInFormer - MLM &47.94$\pm$13.50 & 56.72$\pm$7.28 & 0.334$\pm$0.074 & 0.091$\pm$0.001 & 0.072$\pm$0.002 & 0.418$\pm$0.006 & 0.280$\pm$0.003 & 0.066$\pm$0.066 & Coordinate free\\
MatInFormer - LPP &\underline{43.01$\pm$11.83} & \underline{43.07$\pm$4.31} & \underline{0.322$\pm$0.085} & \underline{0.087$\pm$0.002} & \textbf{0.069$\pm$0.003} & 0.409$\pm$0.007 & 0.267$\pm$0.002 & \textbf{0.064$\pm$0.000} & Coordinate free\\
MatInFormer - LPP(OCP) &43.07$\pm$12.99 & 43.18$\pm$5.27 & 0.338$\pm$0.071 & \textbf{0.086$\pm$0.001} & \underline{0.070$\pm$0.004} & 0.409$\pm$0.007 & 0.300$\pm$0.008 & 0.067$\pm$0.001 & Coordinate free\\
MatInFormer MLM + LPP &44.02$\pm$11.05 & \textbf{42.57$\pm$5.18} & \textbf{0.317$\pm$0.088} & 0.087$\pm$0.002 & 0.070$\pm$0.004 & 0.427$\pm$0.010 & 0.265$\pm$0.002 & \underline{0.065$\pm$0.001} & Coordinate free\\
    Wrenformer   & \textbf{39.63} & 92.33 & 0.3583 & 0.1070 & 0.0811 & \textbf{0.3350} & 0.2986 & 0.0694 & Coordinate free\\
    \midrule
    Roost  & 44.64$\pm$11.73 & 54.39$\pm$4.73 & 0.325$\pm$0.078 & 0.103$\pm$0.002 & 0.080$\pm$0.004 & \underline{0.403$\pm$0.008} & \underline{0.257$\pm$0.006} & 0.085$\pm$0.002 & Structure agnostic\\
    Finder & 47.96$\pm$11.67    & 46.57$\pm$3.74  & 0.320$\pm$0.081 & 0.100$\pm$0.002 & 0.076$\pm$0.003 & 0.645$\pm$0.017  & \textbf{0.231$\pm$0.003}  & 0.084$\pm$0.001 & Structure agnostic\\
    \bottomrule
    
    \end{tabular}
    }
    \caption{MAE of structure-based GNN, coordinate-free transformer and structure agnostic GNN of Matbench materials property prediction. The best performing result among the coordinate free and structure agnostic approaches has been shown in boldface and next best performing result has been underlined. We have rounded off the results to the third significant digit after the decimal wherever necessary.}
    \label{matbench}
\end{table}

\subsection{Effect of pretraining dataset size}

The size of the pretraining data directly influences the performance on downstream tasks. We evaluated the downstream performance using two distinct pretraining datasets. The first dataset contains 642,459 samples, while the second, larger one comprises 997,528 samples. The added difference in the second dataset is the inclusion of OCP\cite{chanussot2021open} data. To understand the impact of dataset size on performance, we pretrained two models using the LPP strategy and assessed their downstream performance. We observed that models pretrained without the OCP data showed better results. As detailed in Table~\ref{matbench}, the LPP model that did not utilize the OCP dataset consistently outperformed the other model. One of the possible reason for this might be that the Open Catalyst data contains unrelaxed structures. These structures, have lattice parameters that are not in line with the relaxed structure crystallography rules and could introduce complexities in feature capture.

\begin{figure}
    \centering
    \includegraphics[width=\textwidth]{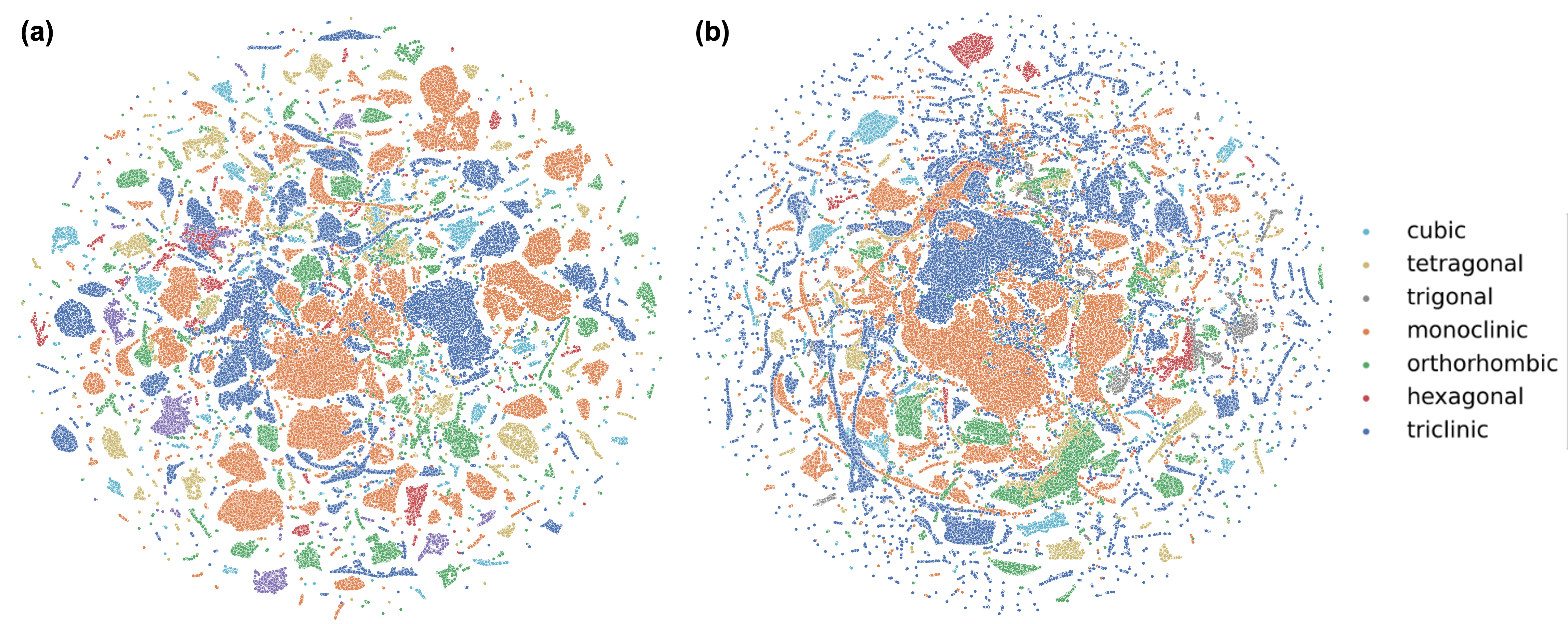}
    \caption{t-SNE plots for two different pretrain datasets}
    \label{tsne}
\end{figure}
\subsection{Exploring the representation space}
To explore the representations learned by the pretrained models, we visualized the representation learned by the pretrained models using lattice parameter prediction strategy. To plot these representations in a 2-D space, we applied t-SNE\cite{van2008visualizing} dimension reduction on the CLS token from the last encoder layer. The t-SNE plots are shown in Figure~\ref{tsne}.

To better understand the patterns in the t-SNE plots, we note that materials primarily cluster based on their crystal system. This suggests that our model can roughly classify crystal structures using just the formula and space group. However, several scattered points appear around the edges of the t-SNE plots. This scattering can be attributed to the inherent difficulty of predicting lattice parameters for models that don't rely on coordinates. The challenge is amplified given that our pretraining datasets range from simple substances, like pure metals, to complex structures like hMOFs, which can contain between 1 to 20 elements and have unit cell volumes that can be hundreds to thousands of times larger.

Comparing two different plots for the pretraining dataset 1 and pretraining dataset 2, we observed that the datasets pre-trained with Open Catalyst Data exhibited more scatter points on the periphery, whereas most of the data clustered in another plot. This could potentially be attributed to the large amount of data from Open Catalyst having the same formula and space group but varying structures, as some of them are distorted structures. This phenomenon is not only evident in the t-SNE plots but also results in inferior performance on the downstream task, as indicated in Table \ref{matbench}. This explanation, however, is one of several possibilities, and the increased scattering observed may also be influenced by other factors.

\subsection{Metal Organic Frameworks(MOFs)}

We assessed the performance of MatInFormer for gas absorption in hMOF using various informatics tokens and compared its results to those of CGCNN, MOFormer, and Stoichiometric-120. To understand the influence of different informatics tokens on the model's efficiency, we conducted ablation studies specifically for hMOF.
The results (MAE) for gas adsorption of hMOF are presented in Table~\ref{hMOF}. MatInFormer, when used without any additional informatics tokens, only performs marginally better than Stoichiometric-120. With the introduction of topology as an informatics token, MatInFormer(T) sees a 24.42\% enhancement in performance, aligning it closely with MOFormer which uses mofid as input. When we incorporate informatics tokens related to unit cell volume and the number of atoms, MatInFormer(TV) boosts its performance by 51.86\%, surpassing CGCNN. Further, when porosity is also considered, MatInFormer(TVP) achieves an improvement of 53.44\% and stands out as the top performer for hMOF gas absorption. We note that the performance of MatInFormer(TVP) is similar to that of MatInFormer(TV).

\begin{figure}[h!]
    \centering
    \includegraphics[width=\linewidth]{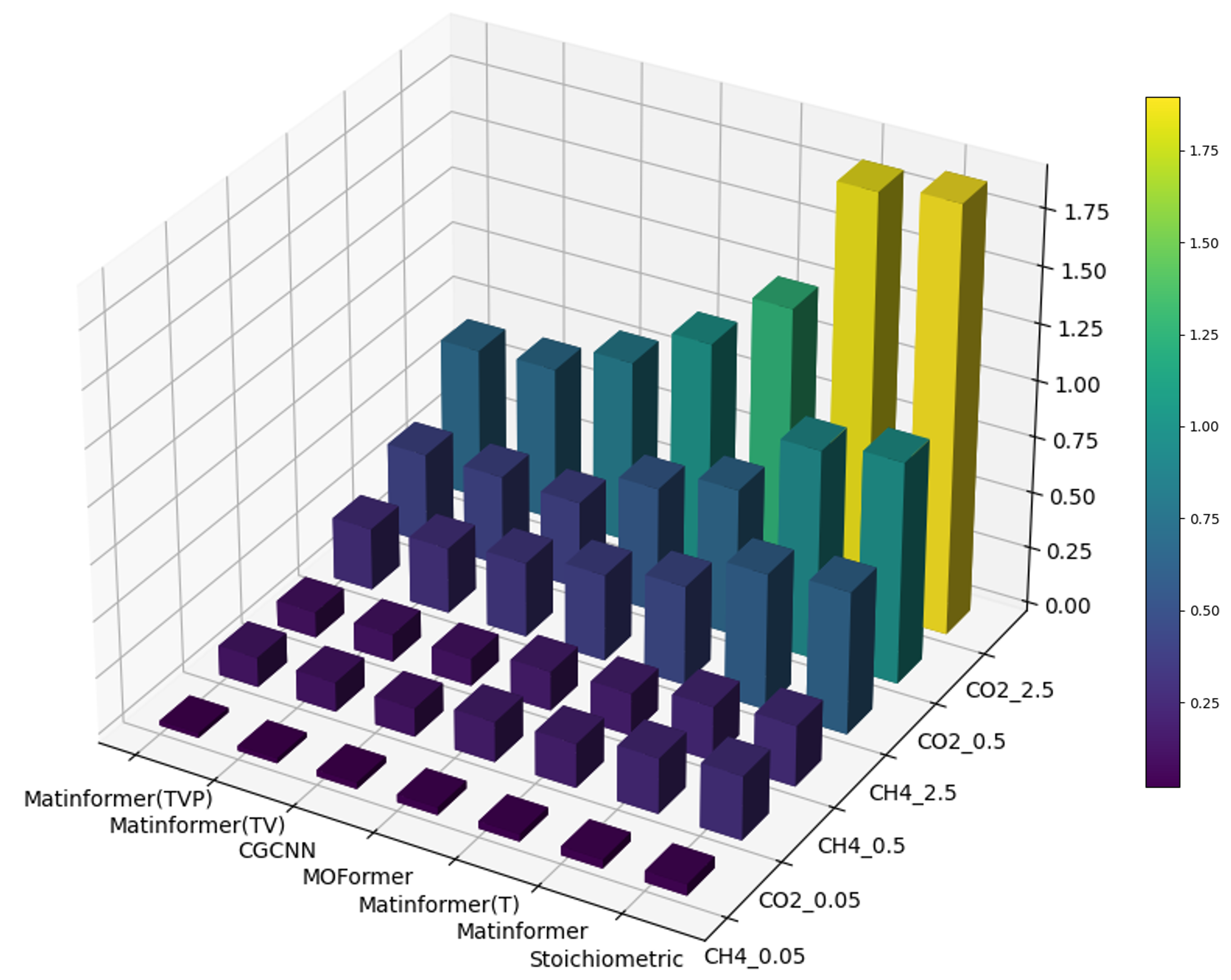}
    \caption{MAE for different models on MOF property prediction tasks. We compare the MatInFormer with models like CGCNN, MOFormer, and Stoichiometric -120. The result for baselines are obtain from MOFormer paper\cite{cao2023moformer}.}
    \label{mof}
\end{figure}

To highlight how MatInFormer compares with other baseline models, we plotted the MAEs of different models based on their average performance in Figure~\ref{mof}. We observe that the performance ranking of the models follows the amount of information they use for making the predictions. In the first tier, both MatInFormer and Stoichiometric-120 mainly use formulas. Since over 95.8\% of hMOF's space group is P1, the formula becomes the main difference. But because most hMOFs have common elements like C, N, O, H, and a few metals, it's hard for these models to distinguish between different MOFs effectively. In the second tier, we have the MatInFormer(T) and which adds topology information helps its performance. Meanwhile, MOFormer does even better because it uses bond information from SMILES in its mofid representation. It's important to note that a lot of the topologies, more than half, are labeled "pcu.cat0" or "pcu.cat1". Even with the help of topology, the model struggles to differentiate more than half of the MOFs. In the third tier, CGCNN uses detailed atomic positions and coordinates, making it really good at telling MOFs apart. But interestingly, MatInFormer versions, specifically MatInFormer(TV) and MatInFormer(TVP), still do better than CGCNN, even though they use less detailed information such as exact atomic coordinates. This shows that MatInFormer is really good at figuring out features like porosity just from cell volume and formula. 


\begin{table}[htb!]
    \centering
     \resizebox{\textwidth}{!}{
    \begin{tabular}{l|lllllll}
    \toprule
    Models & $CO_2$ 0.05bar & $CO_2$ 0.5bar & $CO_2$ 2.5bar & $CH_4$ 0.05bar & $CH_4$ 0.5bar & $CH_4$ 2.5bar \\
    \midrule
    CGCNN              & \textbf{0.126$\pm$0.005} & \textbf{0.391$\pm$0.017} & 0.818$\pm$0.050 & 0.028$\pm$0.001 & 0.121$\pm$0.006  & 0.333$\pm$0.017\\
    MOFormer           & 0.178$\pm$0.002 & 0.558$\pm$0.001 & 1.000$\pm$0.013 & 0.034$\pm$0.000 & 0.174$\pm$0.002 & 0.385$\pm$0.003\\
    Stoichiometric-120 & 0.282$\pm$0.002 & 0.983$\pm$0.005 & 1.895$\pm$0.003 & 0.05$\pm$0.001  & 0.269$\pm$0.001 & 0.631$\pm$0.002\\
    MatInFormer        & 0.245$\pm$0.003 & 0.933$\pm$0.006 & 1.848$\pm$0.005 & 0.038$\pm$0.001 & 0.238$\pm$0.004 & 0.599$\pm$0.006\\
    MatInFormer(T)     & 0.194$\pm$0.001 & 0.661$\pm$0.007 & 1.261$\pm$0.007 & 0.032$\pm$0.001 & 0.186$\pm$0.002 & 0.443$\pm$0.005\\
    MatInFormer(TV)    & \underline{0.129$\pm$0.001} & 0.403$\pm$0.009 & \underline{0.691$\pm$0.007} & \underline{0.022$\pm$0.000} & \underline{0.120$\pm$0.003} & \underline{0.290$\pm$0.005}\\
    MatInFormer(TVP)   & 0.131$\pm$0.002 & \underline{0.398$\pm$0.007} & \textbf{0.672$\pm$0.007} & \textbf{0.021$\pm$0.000} & \textbf{0.113$\pm$0.002} & \textbf{0.270$\pm$0.005}\\
    \bottomrule  
    \end{tabular}
    }
    \caption{MAE values of hMOF absorption. MatInFormer(T) denotes topology as an informatics token, MatInFormer(V) denotes unit cell volume and number of atoms as an informatics token, MatInFormer(P) denotes porosity as an informatics token.}
    \label{hMOF}
\end{table}

\subsection{Visualizing the Attention}
\begin{figure}[htb!]
    \centering
    \includegraphics[width=0.8\linewidth]{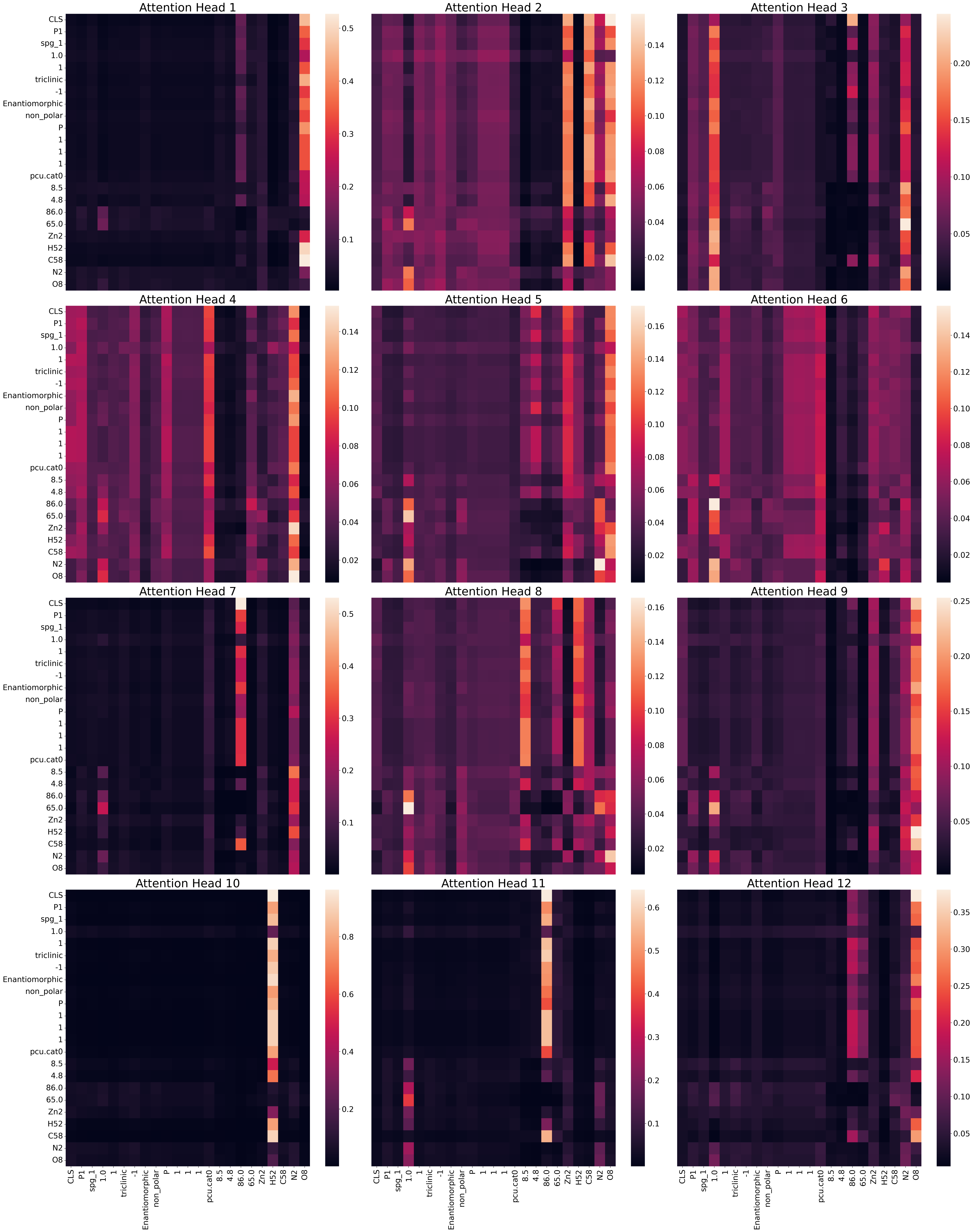}
    \caption{The attention weights for the last encoder layer of the model. Visualizing the attention weights gives a degree of interpretability to the MatInFormer model.}
    \label{last_layer}
\end{figure}
To gain further insights into our model, we analyzed the attention patterns by visualizing attention scores for a randomly chosen example, hMOF-5035957. These scores, shown as a heatmap in \ref{last_layer}, indicate the correlations between tokens for the last layer of the encoder. We observe that some tokens have particularly strong attention signals. Heads 1, 10, and 12, for instance, heavily focus on the formula tokens. Moreover, heads 7 and 11 seem to focus to porosity fractions. Additionally, some of the heads distribute their attention across the space group tokens or evenly spread their attention across all tokens, capturing a more generalized representation of the material. For the downstream tasks, the regression head is coupled with output of the [CLS] tokens. We further analyzed the attention weights of the [CLS] token, the detailed visualization is shown in Figure S6.

\section{Conclusion}

In this study, we developed MatInformer, a coordinate-free transformer framework designed for materials property prediction. MatInformer achieves excellent performance on many different property prediction tasks even without explictly considering structural information. Another advantage of our model lies in its ability to maintain consistent memory use, regardless of the growth in crystal size, similar to other structure-agnostic models. The pretraining strategies that we devised such as predicting lattice parameters, enables the model to gain insights into crystallography, leading to marked improvements in downstream tasks. When applied to hMOF, we investigate  the influence of different features through an ablation study, considering informatics tokens like topology, unit cell volume, and porosity. The use of informatics tokens for MOFs further demonstrates the flexibilty and potential of the MatInFormer to incorporate material information through simple tokenization. Moreover, the visualization of attention heatmaps provides a degree of interpretability for the model. This factor of interpretability might offer insights into how deep learning models generate material representations. 

Although the MatInFormer shows promising results there are multiple challenges ahead. The Transformer model inherently will have difficulties in tokenizing numerical values such as atomic distances and angles, which are crucial for structure-based regression tasks. There is a potential to develop materials specific tokenization strategies. Additionally, multimodal approaches could possibly be leveraged to maximize the benefits inherent to both structure based and structure agnostic models. Overall, we believe MatInFormer is a promising step toward harnessing the power of LLMs to facilitate material discovery.

\section{Data availability}
All Matbench data are available on \href{https://matbench.materialsproject.org}{Matbench Website} and Pretrain data through \href{https://jarvis-tools.readthedocs.io/en/master/databases.html}{JARVIS Databases}
\section{Code availability}
The python code of this work can be found on Github https://github.com/hongshuh/MatInFormer
\bibliography{reference}

\newpage

\setcounter{table}{0}
\renewcommand{\thetable}{S\arabic{table}}
\setcounter{figure}{0}
\renewcommand{\thefigure}{S\arabic{figure}}
\setcounter{equation}{0}
\renewcommand{\theequation}{S\arabic{equation}}
\maketitle
\section{Supplementary Information}
\section{Pretraining Dataset}
We used data from the JARVIS-Tools Database\cite{choudhary2020joint} for MatInFormer's pretraining. This database includes the Open Catalyst dataset\cite{chanussot2021open} and others. Details on these datasets and their sizes are in Table~\ref{pretrain}.

For pretraining dataset 1, we took data from five JARVIS-Tools datasets namely dft\_3d, qe\_tb, alignn\_ff\_db, hMOF\cite{wilmer2012large} and cod\cite{Downs2003},. We combined all the materials from these datasets and removed any duplicate materials.  When calculating lattice parameters with JARVIS tools\cite{choudhary2020joint}, we take the mean value of the repeat data. This was done because we need the lattice parameters for lattice parameter prediction(LPP) strategy.

\begin{figure}
    \centering
    \includegraphics[width = 0.92\linewidth]{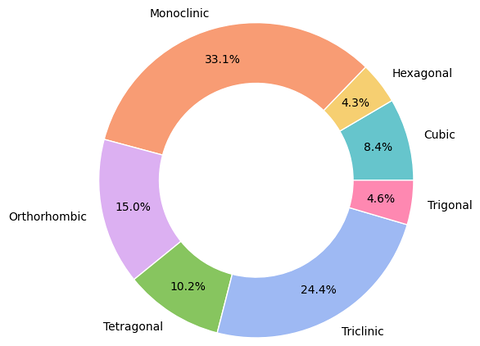}
    \caption{Crystal System distribution for Pretraining Dataset 1 }
    \label{crystal_642}
\end{figure}

\begin{figure}
    \centering
    \includegraphics[width=0.97\linewidth]{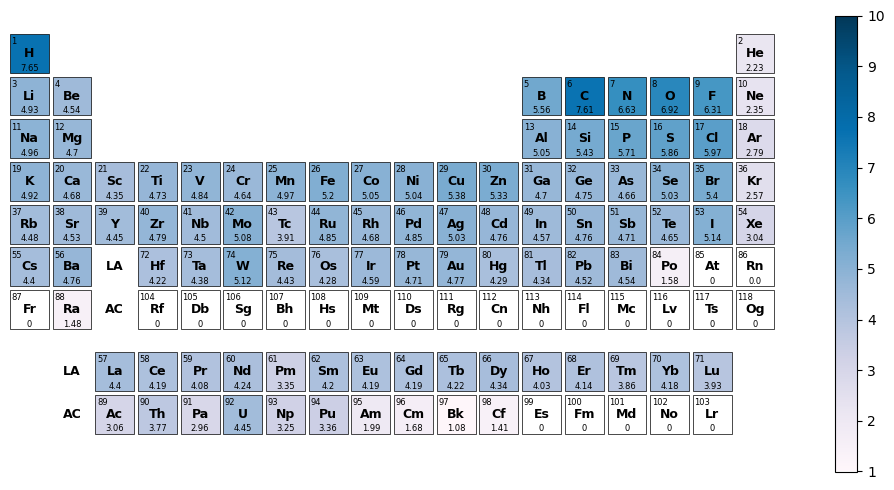}
    \caption{Element distribution for Pretraining Dataset 1 }
    \label{element_642}
\end{figure}

\begin{figure}
    \centering
    \includegraphics[width=0.92\linewidth]{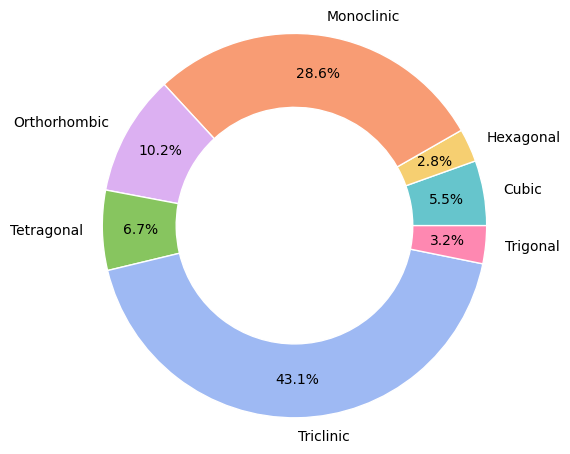}
    \caption{Crystal System distribution of Pretraining Dataset 2}
    \label{crystal_997}
\end{figure}

\begin{figure}
    \centering
    \includegraphics[width=0.95\linewidth]{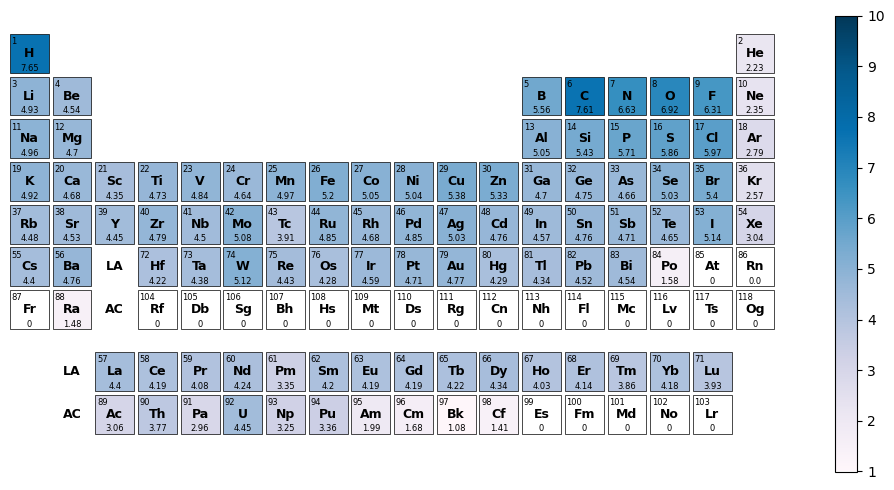}
    \caption{Element distribution of Pretraining Dataset 2}
    \label{element_997}
\end{figure}

Similarly, we developed pretraining dataset 2, by combining all the datasets from Table~\ref{pretrain} and removing any duplicate materials. To explore the pretraining datasets in greater details we plot the crystal system distribution and the elemental distribution. We observe a difference between pretraining dataset 1 and pretraining dataset 2  both in terms of elemental and crystal system distribution. For the pretraining dataset 1 the crystal system and elemental distribution is shown in Figure~\ref{crystal_642} and Figure~\ref{element_642}. For the pretraining dataset 2 the  crystal system and elemental distribution is shown in Figure~\ref{crystal_997} and Figure~\ref{element_997}.  

\begin{table}[]
    \centering
    \begin{tabular}{c|c|c}
    \toprule
    Source & Dataset Name  & Size \\
    \midrule
    JARVIS - Tools & DFT3D & 75993 \\
    JARVIS - Tools & QETB &829574\\
    JARVIS - Tools & alignn ff db & 307113\\
    JARVIS - Tools & hMOF & 137651\\
    JARVIS - Tools & COD & 431778\\
    JARVIS - Tools & Open Catalyst & 510214 \\
    \midrule
     & Pretraining Dataset 1 & 642459\\
     & Pretraining Dataset 2 & 997528
    \end{tabular}
    \caption{Pretraining datasets composition and Size. We use the large dataset repository JARVIS - Tools compile the pretraining datasets in this work.}
    \label{pretrain}
\end{table}

\newpage
\section{Hyperparameters}

\subsection{MatInFormer Hyperparameters}
We used the Roberta\cite{liu2019roberta}  model as the architecture for the MatInFormer. The hyperparameters are detailed in Supplementary Table \ref{Model}.

\begin{table}[]
    \centering
    \begin{tabular}{l|l}
    \toprule
    Hyperparameters & Value\\
    \midrule
    Layers     & 8\\
    Heads number& 12\\
    Attention Dropout & 0.1\\
    Hidden Dropout & 0.1\\
    Linear Layer & 2\\
    Linear Dropout & 0.1\\
    Activation Function & SiLU\\
    Hidden Size & 768\\
    \bottomrule
    \end{tabular}
    \caption{Hyperparameters for the baseline Roberta Model which is used by the MatInFormer}
    \label{Model}
\end{table}

\subsection{Pretraining Hyperparameters}
Different pretraining strategies required various hyperparameters, as displayed in Supplementary Table \ref{pretrain parameter}. Given that Lattice Parameter Prediction (LPP) is more intricate than Masked Language Modeling (MLM), we used distinct hyperparameters for each. However, for the combined MLM + LPP strategy, we employed the same hyperparameters as those used for LPP. 

We use the AdamW\cite{loshchilov2018decoupled} optimizer with a learning rate of 1e-6 and weight decay of 0 for all pretraining strategies. The model was trained with a batch size of 64 for 200 epochs. A CosineAnnealingLR\cite{loshchilov2016sgdr} scheduler with a warm-up is used to stabilize the training process. The learning rate was warmed up during the first 5 \% of the total epoch and then decayed to zero as a cosine function for the rest of the epochs. The model occupied around 5GB memory and it takes 1 day to pretrain on an A6000 GPU.

\begin{table}[]
    \centering
    \resizebox{\linewidth}{!}{
    \begin{tabular}{l|lllllll}
    \toprule
    Method& Dataset & Epoch & Batch Size & Learning Rate & Weight Decay & Warm Up Ratio & Masking Ratio\\
    \midrule
    MLM & 642K & 50 & 64 & 1e-6 & 0.0 & 0.05 & 0.25\\
    LPP & 997K & 150 & 64 & 1e-6 & 0.0 & 0.05 & 0 \\
    LPP + MLM & 642K & 150 & 64 & 1e-6 & 0.0 & 0.05 & 0.25\\
    \bottomrule
    \end{tabular}}
    \caption{Pretrain Hyperparameter, LPP is lattice parameters prediction. LPP with mask is lattice parameters prediction with random masked space group tokens.}
    \label{pretrain parameter}
\end{table}

\subsection{Finetuning Hyperparameters}
For finetuning, the transformer encoder was loaded with the pretrained weights designed for downstream tasks. The hyperparameters used during this finetuning are detailed in Supplementary Table \ref{finetune}. For all models, we employed the AdamW\cite{loshchilov2018decoupled} optimizer and the CosineAnnealingLR\cite{loshchilov2016sgdr} scheduler. During pretraining, we utilized the Mean Square Error (MSE) loss for LPP and LPP + MLM, and cross-entropy for MLM. For the finetuning stage, the Mean Absolute Error (MAE) loss was employed. We would like to note that the hyperparameters during finetuning were optimized using a random search method.

\begin{table}[]
    \centering
    \begin{tabular}{l|llll}
    \toprule
    Dataset &   Batch Size & Learning Rate & Weight Decay & Epoch\\
    \midrule
    JDFT2D (JDFT)\cite{choudhary2017high} & 128 & 1e-5& 0.001 & 50\\
    Phonons\cite{petretto2018high} & 128 & 1e-4 & 0.001&100\\
    Dielectric\cite{petousis2017high} & 128 & 1e-4 &0.001& 100\\
    GVRH\cite{ward2018matminer,deJong2015} & 128 & 1e-4 &0.01& 200\\
    KVRH\cite{deJong2015} & 128 & 1e-4 &0.001& 100\\
    Perovskites\cite{Castelli} &  128 & 1e-5 &0.01& 50\\
    MP-Gap (MP-BG)\cite{materialsproject} & 512 & 1e-4 & 0.0001&200\\
    MP-E-Form (MP-FE)\cite{materialsproject} & 128 & 1e-4 & 0.0001&200 \\
    hMOF \cite{wilmer2012large}& 128 & 1e-5 & 0.01&200\\ 
    \bottomrule
    \end{tabular}
    \caption{Finetuning Hyperparamter for the MatInFormer model. We used the same hyperparameter for finetuning all the different pretrained models. The hyperparmeters for MLM if different have been indicated in (brackets) next to the same hyperparameter for the other finetuned models.}
    \label{finetune}
\end{table}

\section{Grid Point Approach for Porosity}
\begin{figure}
    \centering
    \includegraphics{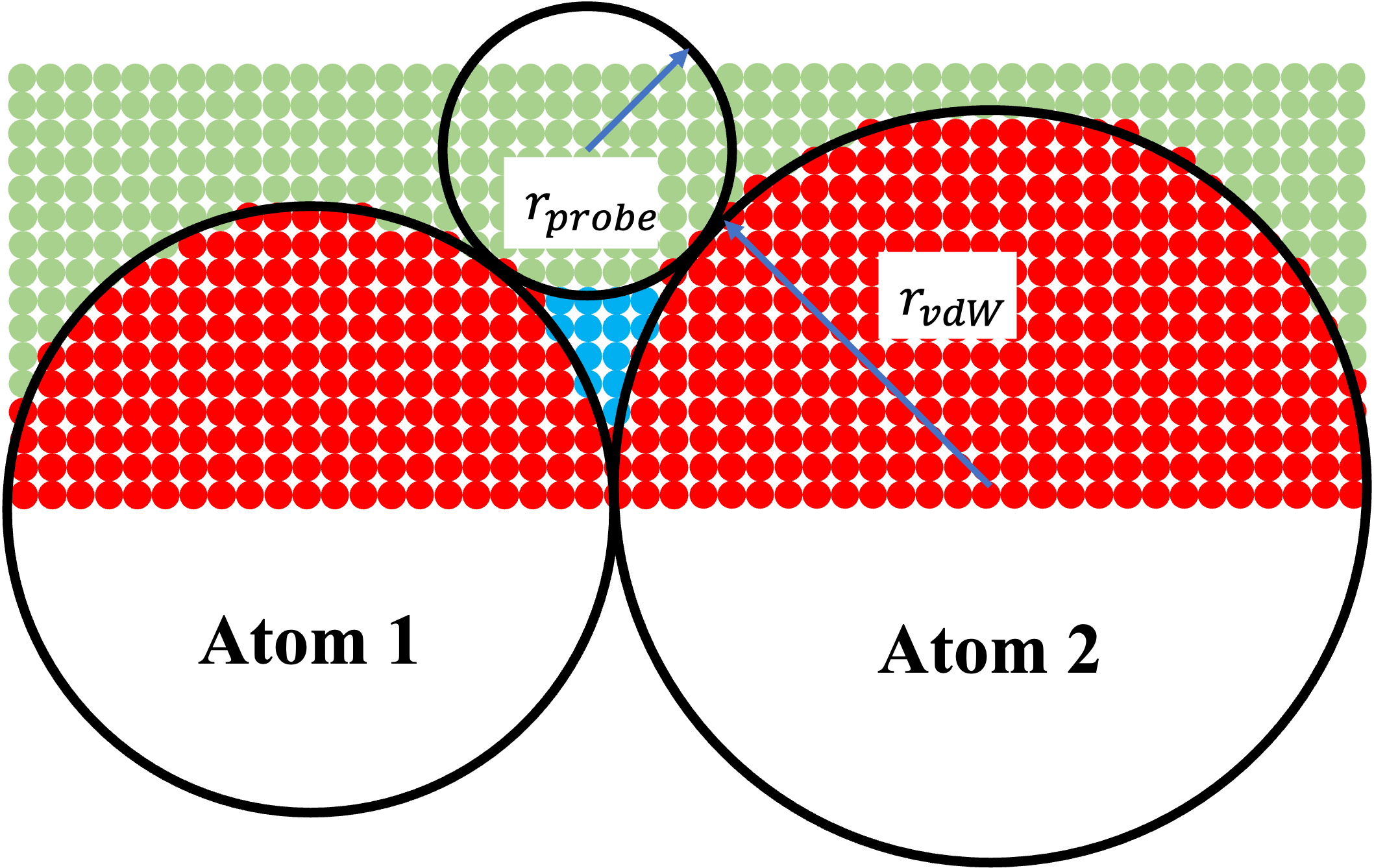}
    \caption{Visual explanation of different grid points in the grid point approach (GPA). Red points inside atoms vdW sphere are considered occupied. Green points are considered unoccupied, Blue points are unoccupied but unaccessible for certain values of $r_{probe}$\cite{trepte2021pore}}
    \label{void_fig}
\end{figure}

We adapt the Grid Point Approach(GPA) from porE\cite{trepte2021pore} for deterministic and systematic analyses of porosities in MOFs. Two properties are calculated, porosity fraction $\Phi_{void}$ of the MOF structure and accessible porosity fraction $\Phi_{acc}$. The visual explanation is shown in Figure~\ref{void_fig}. The porosity is depends on the grid density and accesible porosity is related to the radius of the probe, for our experiment, we use the recommendation setting grid density $\rho_{grid}=5$ and radius of the probe is set to $r_{probe} = 1.2$.

\begin{equation}
\Phi_{void} = \frac{N_{unoccupied}}{N_{total}} \cdot 100 \%
\label{void}
\end{equation}

\section{Hybrid Organic-Inorganic Perovskite(HOIP) result}
In our study on the HOIP dataset, we adopted a 0.6/0.2/0.2 train/validation/test split ratio, with the results clearly outlined in Table \ref{hoip}. We incorporated 16 organic cations as informatics tokens in MatInFormer(O) and observed a marginal performance enhancement of 3.5\%. This improvement is notably superior to that of structure-based GNNs, such as CGCNN. As highlighted in our results and discussion section, the position of atoms emerges as a significantly more crucial feature than the space group. Furthermore, a distinct advantage of our language model is its flexibility; it can easily integrate more informatics tokens. In contrast, GNNs encounter considerable challenges when attempting to encode additional features.
\begin{table}[]
    \centering
    \begin{tabular}{l|l}
    \toprule
    Model     & MAE \\
    \midrule
    CGCNN     & 0.170 $\pm$0.013\\
    MatInFormer & 0.336 $\pm$0.002\\
    MatInFormer(O) & 0.324 $\pm$0.001\\
    \bottomrule
    \end{tabular}
    \caption{Result of HOIP}
    \label{hoip}
\end{table}

\section{Attention Visualization of the CLS token}

\begin{figure}[H]
    \centering
    \includegraphics[width=\linewidth]{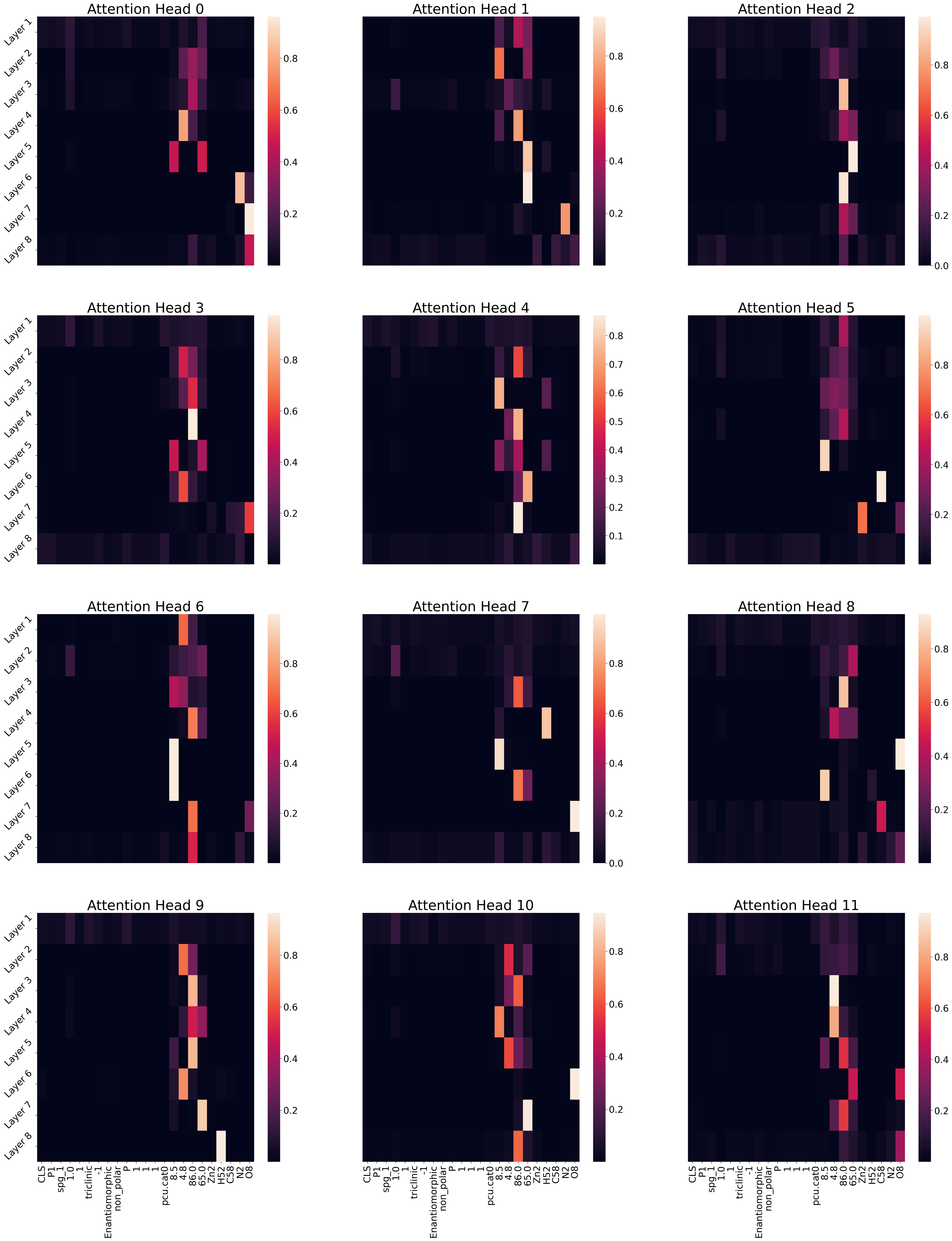}
    \caption{CLS Attention}
    \label{cls}
\end{figure}

In Figure~\ref{cls}, we visualized the attention weights of the [CLS] tokens across different layers. We observe that attention weights focus on informatics tokens, such as topology, volume, and porosity. On the other hand there is negligible attention towards the space group tokens. This can be attributed to the fact that the majority of space groups for hMOF are "P1", making them less relevant for the model to distinguish between hMOFs. Instead, it is the informatics tokens, particularly topology, unit cell volume, and the number of atoms, that play a pivotal role in determining the property. This observation about attention weights aligns with our finetuning results presented in Table 4.

\newpage

\end{document}